\documentclass[sigconf]{acmart}

\renewcommand\footnotetextcopyrightpermission[1]{}  
\AtBeginDocument{%
  }

\usepackage{graphicx}

\usepackage{tikz}
\usetikzlibrary{arrows.meta}

\begin{document}

\title{On the LLM Verification for Argument Graph Generation}

\author{Jakub Bąba}
\email{jakub.baba.stud@pw.edu.pl}
\orcid{0009-0009-7000-4887}
\affiliation{%
  \department{Institute of Computer Science}
  \institution{Warsaw University of Technology}
  \city{Warsaw}
  \country{Poland}}

\author{Jarosław A. Chudziak}
\email{jaroslaw.chudziak@pw.edu.pl}
\orcid{0000-0003-4534-8652}
\affiliation{%
  \department{Institute of Computer Science}
  \institution{Warsaw University of Technology}
  \city{Warsaw}
  \country{Poland}}

\renewcommand{\shortauthors}{J. Bąba and J.A. Chudziak}

\begin{abstract}
  Multi-agent LLM systems can formalize natural-language argumentation end-to-end, turning plain text into structured graphs by classifying components, connecting them with relations, and enriching the resulting graphs into formal models. Quality assurance for those graphs is almost always another LLMs: a reviewer, judge, or critic agent. On the other side, in adjacent neuro-symbolic fields, such generated output is often verified symbolically by external checks and solvers. In this work, we argue that in argument graph verification, the allocation should be similar, and based on what each instrument can guarantee. We propose layered verification: starting from deterministic checks, moving to solver-based and LLM reviews, and utilizing debate only under uncertainty, leaving human judgment for what no other instrument can decide. We ground our principle with a study on human-annotated graphs with injected structural defects. An LLM reviewer detects them at near-ceiling recall, including cycles in the support relation, but it objects to almost one in a ten correct graphs, even though it should not, happens to be inconsistent in verdict on identical inputs and is significantly slower than the equivalent exact check. It shows that while LLM reviewer does not lack the ability to detect structural defects, it is worth having a deterministic verification gates.
\end{abstract}





\maketitle

\section{Introduction}
\label{sec:intro}

Argument Mining (AM) is one of the fields that has benefited from the development of Large Language Models (LLMs). They are used to the argument graph generation from natural text, with the subtasks from classifying components and relations in established annotation schemes~\cite{stab-gurevych-2017-parsing} to constructing structured argument graphs~\cite{freedman2025claimverification,gorur-etal-2025-large}. While fine-tuned models define the supervised state of the art~\cite{cabessa-etal-2025-argument}, training-free approaches have shown competitive results across AM subtasks~\cite{li2025largelanguagemodelsargument}, and recent multi-agent solutions, including debate between agents defending competing labels~\cite{baba2026aric} and critic refinement loops~\cite{baba2026cafgen}, further improve the reliability of such formalization without task-specific training. These systems fit into the broader trend of multi-agent quality mechanisms for LLM reasoning~\cite{du2024multiagent,pitre-etal-2025-consensagent}.

A generated graph is a candidate and should be verified whether is it correct and should be kept. We call this decision verification: a check that can reject or flag a candidate graph. Currently, this quality assurance is based almost entirely on LLM instruments, such as reviewer agents, judge agents, or multi-agent debate itself. However, LLM graph reasoning degrades as the graphs grow, with the cycle detection as one of the measured tasks~\cite{nlgraph}, self-verification from feedback alone has shown to not always succeed~\cite{kamoi-etal-2024-llms,huang2024largelanguagemodelsselfcorrect}, and verification failures recur as a documented failure
mode across multi-agent LLM systems~\cite{cemri2026why}. Such evidence shows that LLMs is not a correct instrument when used solely.

By comparison, in adjacent neuro-symbolic fields, generated output is usually checked by exact instruments. In formal theorem proving, candidate proofs are accepted only when a proof checker verifies them~\cite{ren2025deepseekproverv2}. In LLM-based logical reasoning, formalized problems are decided by deterministic solvers, whose error messages additionally drive the refinement of the formalization~\cite{pan-etal-2023-logic,olausson-etal-2023-linc}. Closer to argumentation, natural-language arguments have been autoformalized and checked for fallacies with SMT solvers~\cite{lalwani-etal-2025-autoformalizing}, and SMT reasoning over LLM-extracted knowledge has been applied to legal reasoning~\cite{sadowski2025legal}. Structural checks are likewise a commodity, though grammar-constrained decoding cannot express the graph-level properties such as acyclity~\cite{dong2025xgrammarflexibleefficientstructured}. Tiered arrangements combining these type of checks have begun to appear: in argument-graph
evaluation, where mechanical checks cover structure
alone~\cite{li2025archenoveltaskevaluate}, and in reasoning-trace verification, where they extend
to arithmetic and to constraints supplied with the problem, leaving deduction
and the validity of the final conclusion to an audit LLM~\cite{zhong2026verytraceverifyingreasoningtraces}.

Argument mining gives us a strong foundation for the same practice, as some areas of NLP produce target structures with a rich formal machinery: acceptability semantics for abstract frameworks~\cite{dung1995}, structured formalisms with established translations between them~\cite{modgil2014,vangijzel2012}, and maintained tooling and benchmarks~\cite{odekerken2023pyarg}. While these symbolic components do appear in LLM-based argumentation systems, they serve as the reasoning mechanism over the generated structures rather than as their verification: gradual semantics computes verdicts from LLM-built frameworks whose base scores and edge polarities are never checked~\cite{freedman2025claimverification}. Currently, there is a lack of research examining how verification effort should be organized in AM pipelines, although this verification would be a natural extension to them.

In this work, we show that verification of generated argument graphs should be allocated by what each instrument can guarantee, rather than what it happens to detect. We propose layered verification, where deterministic validators run on every candidate graph, solver-based semantics checks on every structurally valid one, and LLM review and debate are used only for what no other exact instrument can decide. We ground our proposal on 112 human-annotated argument graphs, finding that an LLM reviewer detects injected structural defects at near-ceiling recall yet objects to graphs that are correct by annotation, changes its verdict on identical input, and cannot check a property the schema never declared.

\section{Problem Positioning}
\label{sec:problem}

A generated argument graph has properties that can be split into three types, and each of them should be verified on its own way.

Structural properties are fixed, based on the schema type and the graph properties. Such properties include required fields, unique identifiers, graph shape (e.g. acyclity and orphaned nodes), statement texts that are actually in the source. They introduce errors like duplicate identifiers, invalid links, missing or malformed fields.
Semantic properties are based on the formalism that the graph is written in. These are properties like whether or not the graph can be evaluated at all or if its main claims survive attacks on it. They are decidable if the target formalism have such semantics.
Subsymbolic properties are the rest of the properties. It can be, but is not limited to: verifying logical links between each of the arguments, correct proof standards, matching the general idea of the text with the graph. For that, no tool can assess it exactly, as it requires contextual and linguistic judgment.

Current pipelines focus vastly on the LLM review on all of the types of properties. While its quality is becoming progressively higher, it is still the least predictable instrument compared to deterministic validators or solvers. The question that we ask in this work is whether this is the correct assessment. The goal is to decide which properties should be decided by which instruments and on what ground these assignments are made.

\section{Approach, Method and Tools}
\label{sec:approach}

The allocation we propose orders instruments by what each can decide and what it costs to run. The allocation principle is not itself new; Section~\ref{sec:intro} noted its independent emergence in reasoning-trace and argument-graph verification. What argumentation contributes is that every level can be filled with instruments that already exist and need no per-instance specification. Deterministic validators are exact and nearly free, so they run on every candidate. Solver-based checks are exact within their formalism and cheap at the scale of a single graph, so they run on every structurally valid candidate. LLM review is fallible and expensive, so it is restricted to properties no exact instrument decides. Debate, the least predictable instrument, is admitted only where the system's own uncertainty warrants it, and human judgment only where debate does not settle the matter.

Each level strains the next: a candidate reaches a costly instrument only after the cheaper ones have passed it, and each instrument checks only the properties for which it is the cheapest sufficient tool. This inverts the allocation described in Section~\ref{sec:problem}, making the absent exact instruments unconditional and the unconditional fallible ones gated.

\begin{figure*}[t]
\centering
\resizebox{\textwidth}{!}{
\begin{tikzpicture}[
  font=\small,
  gate/.style={rounded corners=1.5pt, draw, line width=0.6pt,
               minimum width=0.3cm, minimum height=2.9cm},
  gname/.style={font=\small\bfseries, align=center, inner sep=1pt},
  gsub/.style={font=\scriptsize, text=black!60, align=center, inner sep=1pt},
  flow/.style={-{Stealth[length=2.2mm]}, black!70, line width=0.7pt},
  drop/.style={densely dashed, black!55, line width=0.5pt},
  seg/.style={fill=black!12}
]
 
\fill[seg] (1.90,-1.00) rectangle (3.40, 1.00);
\fill[seg] (3.40,-0.70) rectangle (5.20, 0.70);
\fill[seg] (5.20,-0.50) rectangle (7.00, 0.50);
\fill[seg] (7.00,-0.35) rectangle (8.80, 0.35);
\fill[seg] (8.80,-0.25) rectangle (10.03,0.25);
 
\draw[flow] (2.05,0) -- (3.15,0);
\draw[flow] (3.70,0) -- (4.95,0);
\draw[flow] (5.50,0) -- (6.75,0);
\draw[flow] (7.30,0) -- (8.55,0);
\draw[flow] (9.10,0) -- (9.90,0);
\draw[flow] (11.85,0) -- (12.55,0);
\node[gsub, anchor=west, text=black] at (12.62,0) {accepted};
 
\node[draw=yellow!45!black, fill=yellow!35, rounded corners=3pt,
      line width=0.6pt, minimum width=1.9cm, minimum height=2.0cm]
      at (0.95,0) {\textbf{Generator}};
\node[gsub] at (2.58,0.42) {candidate\\graph};
 
\node[gate, fill=blue!15,   draw=blue!60!black]   at (3.40,0) {};
\node[gate, fill=blue!15,   draw=blue!60!black]   at (5.20,0) {};
\node[gate, fill=orange!22, draw=orange!65!black] at (7.00,0) {};
\node[gate, fill=orange!22, draw=orange!65!black] at (8.80,0) {};
 
\node[gname] at (3.40,2.85) {Deterministic validators};
\node[gsub]  at (3.40,2.55) {always, exact};
\draw[black!40, line width=0.4pt] (3.40,2.36) -- (3.40,1.52);
 
\node[gname] at (5.20,1.95) {Solver checks};
\node[gsub]  at (5.20,1.65) {formal semantics};
 
\node[gname] at (7.00,2.85) {LLM review};
\node[gsub]  at (7.00,2.55) {subsymbolic only};
\draw[black!40, line width=0.4pt] (7.00,2.36) -- (7.00,1.52);
 
\node[gname] at (8.80,1.95) {Debate};
\node[gsub]  at (8.80,1.65) {below $\tau$ only};
 
\node[gsub] at (9.48,0.62) {contested\\residue};
 
\node[draw=violet!60!black, fill=violet!12, rounded corners=3pt,
      line width=0.6pt, minimum width=1.75cm, minimum height=1.5cm,
      align=center] at (10.90,0) {\textbf{Human}\\\textbf{judgment}};
 
\draw[drop] (3.40,-1.50) -- (3.40,-2.20);
\draw[drop] (5.20,-1.50) -- (5.20,-2.20);
\draw[drop] (7.00,-1.50) -- (7.00,-2.20);
\draw[drop] (8.80,-1.50) -- (8.80,-2.20);
\draw[drop] (10.90,-0.80) -- (10.90,-2.20);
\draw[drop, -{Stealth[length=2.2mm]}]
      (10.90,-2.20) -- (0.95,-2.20) -- (0.95,-1.05);
\node[gsub] at (6.00,-2.45) {reject $\rightarrow$ repair};
 
\end{tikzpicture}
}
\caption{Verification organized as levels of increasing cost and decreasing reliability.}
\label{fig:cascade}
\end{figure*}
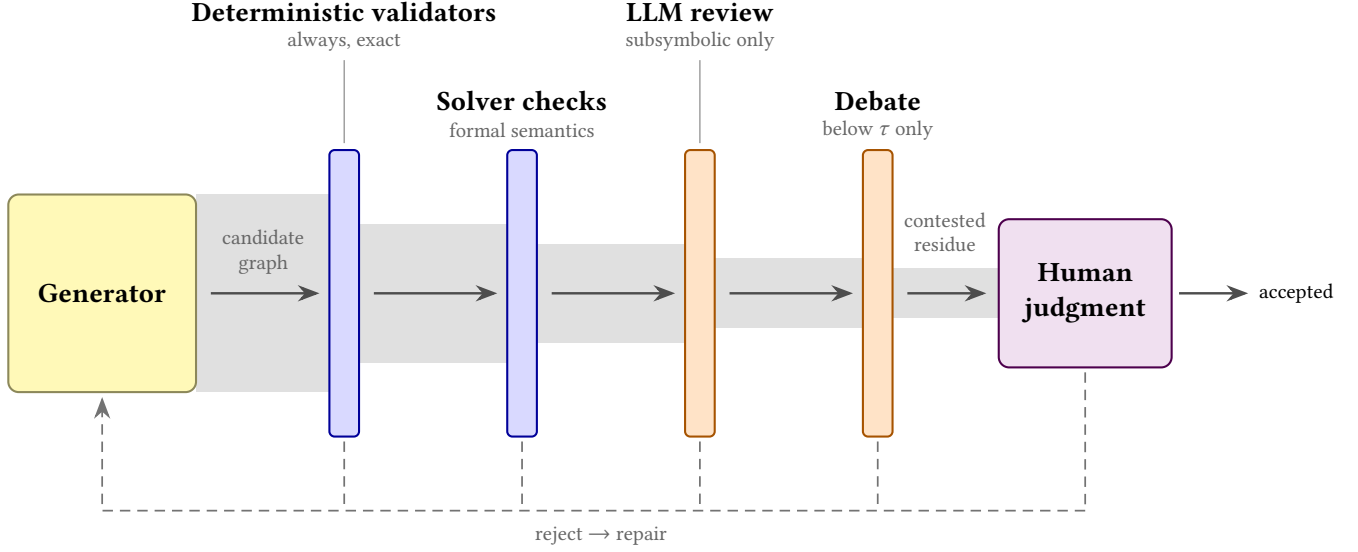

\paragraph{Deterministic validators.} The first level decides syntactic properties exactly and at negligible cost. For enrichment it consists of the three checks: schema validation, verifying required fields and defined enumerations; graph traversal, identifying duplicate identifiers, dangling references, orphans and cycles; and span matching, confirming that every statement text occurs in the source. Violations at this level are the most important, immediately blocking the candidate. For each violation there is reported  an exact location, so the generator receives a precise repair report.

\paragraph{Solver-based checks.} The second level checks are what current pipelines don't verify: properties following from the formal semantics of the generated graph. In order to decide them, we translate the enriched model into ASPIC+ via the van Gijzel-Prakken construction~\cite{vangijzel2012} and evaluate the induced Dung framework with maintained tooling~\cite{odekerken2023pyarg}. Grounded semantics are sufficient for this purpose, since the translation to them guarantees that the standard semantics coincide. A solver appear in argumentation systems already, but as the reasoning engine that computes verdicts from a graph: in that way, it is a inference over a graph rather than check on it, and it verifies nothing. Here, the same instrument is placed as a gate. A graph that cannot be evaluated is blocked. If it can be evaluated, but its own main claim comes out undecided or defeated, the graph is flagged, even if its local features are plausible. While this isn't a definite rejection, it might be a valuable attachment to the next level of LLM review evaluating this concern.

\paragraph{LLM review and debate.} What survives is subsymbolic. Scheme selection, proof standards and semantic coherence accounted for roughly half the flags in~\cite{baba2026cafgen}, and there an LLM reviewer is the appropriate instrument. Its scope narrows: it receives a graph already known to be structurally valid, together with the solver's findings, so its judgment concentrates on whether the graph matches the text. Debate serves within this level, but only under the uncertainty gate, since~\cite{baba2026aric} reports benefit only below a confidence threshold; applied to confident predictions it introduces doubt without improving accuracy. Cases still unresolved are left to human assessment.

\section{Experiments and Evaluation}
\label{sec:evidence}

Our method rests on three claims: that an LLM reviewer cannot serve as a verification gate however high its recall, that a deterministic validator supplies over the same defect classes the guarantees it lacks, and that a property the schema never declares is checked by nothing at all. We ground these claims on the experiments: the first two on human-annotated argument graphs with injected structural defects, and the third on graphs produced under the same schema by a generator.
 
\subsection{Experimental Setup}
\label{sec:setup}
 
We use arg-microtexts~\cite{Peldszus2015AnAC}, 112 texts annotated with complete argument graphs, converted deterministically into a Carneades-style schema. The corpus provides gold structure, verbatim text spans, a single root claim per text, and tree-shaped graphs, so a seed is correct by annotation and any injected cycle is genuinely injected. Its 63 annotated undercutters map onto Carneades exception premises, since an accepted exception blocks its argument. Conversion yields 576 statements and 385 arguments, all accepted by the validator. Schemes and proof standards are not annotated in the corpus and are assigned mechanically; this is sound for the defect classes below, which turn on graph structure and taxonomy membership rather than on the appropriateness of a label.
 
We compare a deterministic validator, performing schema validation, graph traversal and span anchoring, against an LLM reviewer (Gemini~3.7~Flash, temperature~0.1) prompted with the role description of a published Reviewer agent~\cite{baba2026cafgen}. The reviewer prompt states the exception convention explicitly.
 
We inject one defect per graph per class: \textbf{D1}, a required field removed; \textbf{D3}, a duplicate identifier; \textbf{D4}, a dangling reference; \textbf{D5}, a cycle in the support relation. The classes differ in the locality of the evidence a violation requires: a single object for D1, a pair for D3 and D4, and the transitive closure of the edge set for D5. Mutants whose validator findings fall outside the intended class are discarded (one of 405), so no mutant is detectable for the wrong reason. Each verifier is queried three times on 404 mutants and on the 112 unmodified graphs, for 1{,}548 calls.
The converter, validator, mutation operators and prompts are released with the paper.
 
\subsection{Defect Detection}
\label{sec:detection}
 
\begin{table}[t]
  \centering
  \small
  \caption{Detection of injected defects by the LLM reviewer.
  \emph{any}/\emph{all}: flagged in at least one / in all of three runs;
  \emph{unst.}: runs disagree. The validator row holds by construction.}
  \label{tab:detection}
  \begin{tabular}{@{}lrrrr@{}}
    \toprule
    Class & $n$ & any & all & unst. \\
    \midrule
    D1 field removed  & 111 & 0.96 & 0.93 & 0.04 \\
    D3 duplicate id   & 111 & 1.00 & 1.00 & 0.00 \\
    D4 dangling ref.  & 112 & 1.00 & 1.00 & 0.00 \\
    D5 support cycle  &  70 & 0.97 & 0.97 & 0.00 \\
    \midrule
    Validator, all    & 404 & 1.00 & 1.00 & 0.00 \\
    \bottomrule
  \end{tabular}
\end{table}
 
Table~\ref{tab:detection} reports detection rates. The LLM reviewer finds injected defects at near-ceiling recall, and recall does not decrease with the locality of the evidence required: cycles in the support relation are detected as reliably as a missing field. Degradation with graph complexity has been reported elsewhere~\cite{nlgraph}. Our graphs contain between three and ten nodes, which we take to be why it does not appear here. Detection therefore does not separate the two instruments.
 
\subsection{Behaviour on Correct Graphs}
\label{sec:guarantees}
 
\begin{table}[t]
  \centering
  \small
  \caption{Behaviour on the 112 graphs that are correct by annotation.
  Validator properties hold by construction; reviewer values are observed
  rates.}
  \label{tab:guarantees}
  \begin{tabular}{@{}lll@{}}
    \toprule
     & Validator & LLM reviewer \\
    \midrule
    Silent passes                   & none      & D1 0.04, D5 0.03 \\
    Objections to correct graphs    & 0.00      & 0.098 \\
    Verdict change, identical input & none      & 0.062 \\
    Median latency                  & ${\sim}1$\,ms & 2{,}758\,ms \\
    \bottomrule
  \end{tabular}
\end{table}
 
Every graph in Table~\ref{tab:guarantees} is correct by annotation, so every structural objection is an error. The reviewer objected to 11 of 112 graphs, and on 6.2\% its verdict differed across three identical calls at temperature~0.1. Objections were more frequent on graphs containing exception premises (13.8\% against 5.6\%), one flag code being \texttt{self\_defeating\_exception}: the reviewer treated a premise that blocks an argument as one that supports it. Because schemes and proof standards in the seeds are assigned mechanically, we count only objections to fields that are gold; of 261 raw objections, 218 concerned schemes and were largely correct complaints about those synthetic values.
 
Higher recall would not repair these properties. In a Creator-Reviewer refinement loop, an unsound and non-deterministic gate returns roughly one correct graph in ten for revision, and does so inconsistently, which is consistent with Reviewer feedback reported as ``debatable or overly pedantic''~\cite{baba2026cafgen}. The validator provides soundness, completeness on its classes and determinism as properties of the check rather than as observed rates, at three orders of magnitude lower latency.
 
\subsection{Unspecified Properties}
\label{sec:unspecified}
 
The third claim concerns properties to which no instrument is assigned. Given the gold structure of each text, the same model produced an enriched graph, which the validator then checked. Statement type, proof standard and argument polarity are declared in our schema as closed enumerations; the argument scheme was declared only as a free string with an informal instruction to use Walton's taxonomy.
 
The declared fields were never violated across 576 statements and 385 arguments. The undeclared field was violated in 27 arguments, affecting 24 of 112 graphs (21\%), with a further 32 non-canonical surface forms of real schemes. Observed values include \texttt{undercutter\_argument}, a defeat type from a different formalism, and \texttt{argument\_from\_convergent\_evidence}, a structure from argument diagramming: the generator fills a scheme slot with a relation label. Neither instrument could detect this, the validator having no vocabulary to compare against and the reviewer no criterion to apply. The class was undecidable not because it is hard, but because the schema never declared the permitted values - the allocation failure a layered arrangement is intended to expose, since each level must be told which properties it owns.

\section{Discussion and Future Work}

Our layered verification approach is a proposal. The measurements come from one corpus, one target schema and one model. Schemes and proof standards in our seeds are assigned mechanically rather than annotated, so the false-positive rate counts only objections to fields that are gold, and the scheme-selection level is argued for rather than measured. The solver-level attributions presented in our method are analytical, predicting what each instrument would decide rather than confirming it based on the experiments. Finally, the exact levels are also exact only up to their implementation: even mature proof checkers have had bugs that accepted invalid proofs~\cite{ren2025deepseekproverv2} - but the idea is to reduce the probability of the failures with each level rather than eliminating failure altogether.

These gaps set the next phases. Implementing the solver level and evaluating the system end-to-end would measure what we can currently only predict: the effect on refinement iterations, acceptance rates and the cost of LLM calls.The exact gating mechanism is something, that could deserve a separate study, as it might depend on the confidence that predicting model was calibrated to; other rules for gating, including decision made by the LLM itself, could be also evaluated. The solver level also constrains the choice of target formalism, since it exists only where generated graphs have an evaluable semantics and maintained tooling, and we consider this as a criterion for deciding what LLM pipelines should generate in the first place. Finally, scheme and proof-standard quality remains hard to measure at corpus scale. A correct and realistic approach would split the work, with solvers checking consistency across the whole corpus and human judgement verifying correctness on a small sample.

\section{Conclusion}

Pipelines for argument graph generation delegate quality assurance almost entirely to further LLMs, although much of what those instruments check is decidable by cheaper, exact tools. This raises a question of which instrument should decide about which property, and what is that assignment based on. We proposed a mechanism of layered verification: deterministic validators on every candidate, solver-based checks on every structurally valid one, and LLM review and debate admitted only under uncertainty, with a note to human reserved for what nothing cheaper can decide.

Our study shows that an LLM reviewer can detect structural defects at near-ceiling recall, but lacks guarantees of the deterministic validators. It objected to one correct graph in ten, changed its verdict on identical input, and came with the vastly higher cost. We've shown that the idea of layered verification is thus needed. Implementing the solver level and evaluating the arrangement end to end would make it possible to test our position and make a step towards trustworthy verifiers in AM. More broadly, argumentation is one of the few areas of natural language processing whose target structures come with semantics to verify against, and its pipelines should use it.

\begin{acks}
The work reported in this paper was supported by the \grantsponsor{ncn}{Polish National Science Centre}{https://ncn.gov.pl} under grant \grantnum{ncn}{2024/06/Y/HS1/00197}.
\end{acks}

\bibliographystyle{ACM-Reference-Format}
\bibliography{bibliography}

\end{document}